\documentclass{article}
\usepackage{amsmath}
\usepackage{amssymb}
\usepackage{graphicx}
\usepackage{epsfig}

\title{Application of Hopfield Network to Saccades}
\author{Teruyoshi Washizawa\\
Faculty of Letters, Keio University}

\begin{document}

\date{}

\maketitle

\begin{abstract}
Human eye movement mechanisms (saccades) are very useful for scene analysis, including object representation and pattern recognition.
In this letter, a Hopfield neural network to emulate saccades is proposed.
The network uses an energy function that includes location and identification tasks.
Computer simulation shows that the network performs those tasks cooperatively.
The result suggests that the network is applicable to shift-invariant pattern recognition.
\end{abstract}

\section{INTRODUCTION}
The human eye is distinguished from commercially available electronic cameras by virtue of having much better resolution in the fovea near the optical axis.
The eccentricity of the retina, which produces a large field of view and high acuity at the fovea, causes the human visual system to have quick jerky eye movements - i.e., saccades.
It was shown that an artificial visual system equipped with saccades could reduce the number of processing units required for pattern recognition \cite{Nakano:Niizuma:Omori1989} and could greatly simplify the calculations in low-level vision systems \cite{Ballard1989}.
The saccades involve two subtasks; i.e., a location task and an identification task.
The location task fixates the location of a pattern in a wide field of view using \it{a priori} knowledge of the pattern class of interest.
On the other hand, the identification task determines the class of a pattern under the assumption that the location of the pattern can be fixated.
These subtasks are therefore cooperative.
Recently, Hopfield and his colleagues have shown that a symmetric interconnected neural network (a Hopfield neural network) can perform error corrections in associative retrieval \cite{Hopfield1982} and lead to appropriate solutions in optimization problems \cite{Tank:Hopfield1986,Takeda:Goodman1986}.
In this letter, we treat the saccades as the optimization problem and apply the Hopfield network to the emulation of saccades.
Computer simulation confirms that the network performs these subtasks cooperatively.
It also suggests that the network is applicable to shift-invariant pattern recognition.

\section{ENERGY FUNCTION FOR SACCADES}
\begin{figure}[hbtp]
  \begin{center}
    \includegraphics[scale=0.7]{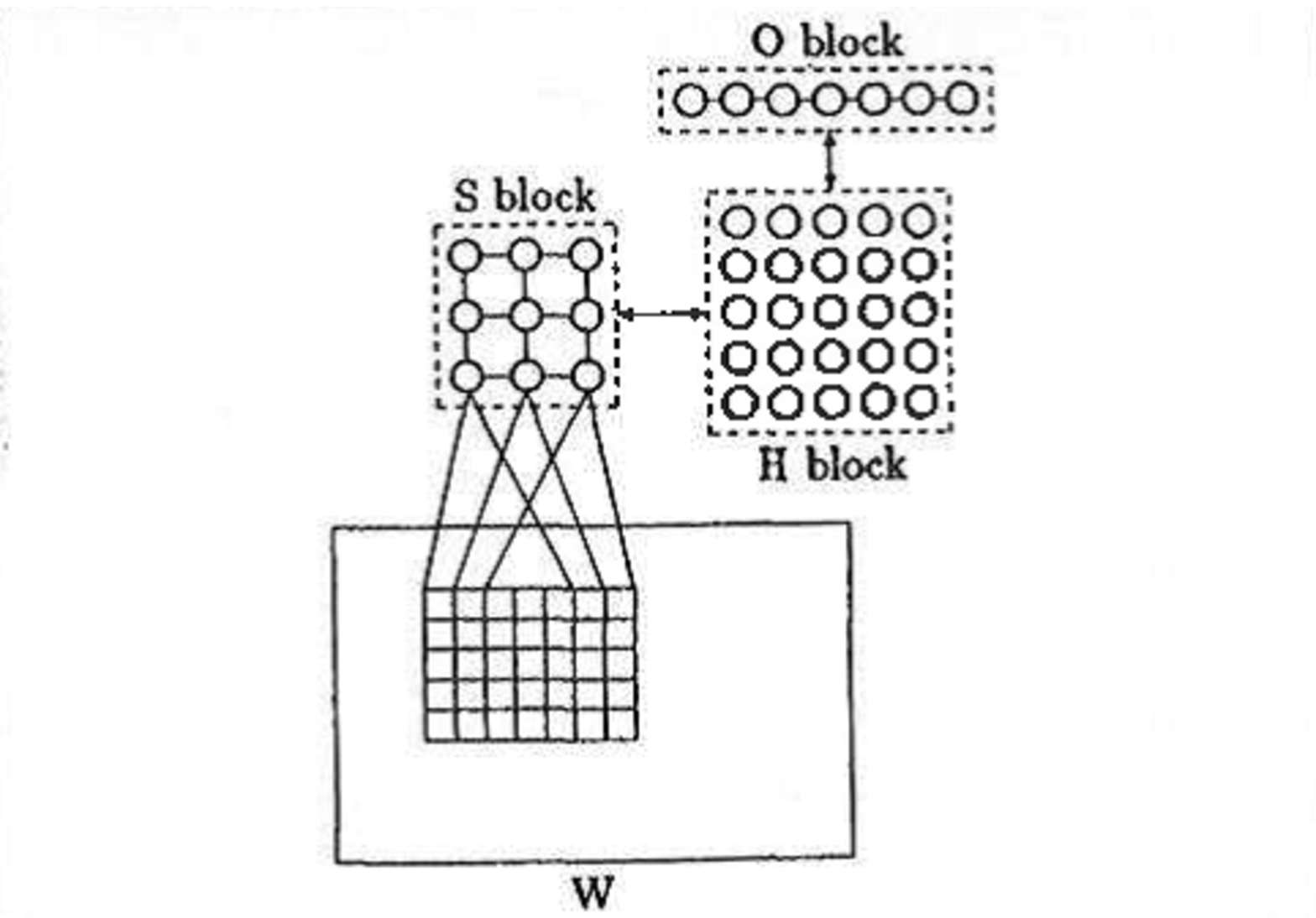}
    \caption{Schematic view of our network. Fully interconnected in the $S$ and $O$ block. Fully interconnected between the $S$ and $D$ block and between the $D$ and $O$ block.}
    \label{figure01}
  \end{center}
\end{figure}

The network (shown in Fig.\ref{figure01}) comprises three blocks of neurons called the saccades (S) block, hidden (H) block, and output(O) block, respectively.
Its energy function is defined as follows:
\begin{eqnarray}
E & = & C_{0} \sum_{i=0}^{I-1} \sum_{j=0}^{J-1} \left(  \sum_{x=0}^{X-1} \sum_{y=0}^{Y-1} Z_{i+x,j+y}^{l,m}V_ 
{x,y}-^{S} - V_{i,j}^{H} \right)^2 \nonumber \\
& + & C_{1} \sum_{i=0}^{I-1} \sum_{j=0}^{J-1} \left( \sum_{n=0}^{N-1} T_{i,j}^{n}V_{n}^{O} - V_{i,j}^{H}  
\right)^2 \nonumber \\
& + & C_{2} \left( \sum_{x=0}^{X-1} \sum_{y=0}^{Y-1} V_{x,y}^{S} - 1 \right)^2 \nonumber \\
& + & C_{3} \left( \sum_{n=0}^{N-1} V_{n}^{O} - 1 \right)^2
\label{EnergyFunction01}
\end{eqnarray}
where
\begin{equation}
Z_{i,j}^{l,m} = A_{l+i-(I+X)/2,m+j-(J+Y)/2} W_{l+i-(I+X)/2, m+j-(J+Y)/2}
\label{EnergyFunction02}
\end{equation}
and
\begin{equation}
A_{l+i-(I+X)/2, m+j-(J+Y)/2} = \left\{
\begin{array}{ll}
1 & \mbox{if $0 \le i < I$ and $0 \le j < J$} \\
0 & \mbox{otherwise}
\end{array}
\right.
\end{equation}
where $(l,m)$ is a gaze position, $W$ a pixel image of an input pattern, $A$ a window function, $u_{i,j}^{*}$ and activity of the $(i,j)$th neuron of $*$ block, $V_{i,j}^{*}$ the output of the $(i,j)$th neuron of $*$ block ($V_{i,j}^{*} = (1 + exp(-u_{i,j}^{*})^{-1}$), $T^{n}$ the $n$th template embedded in the weights between the $D$ and $O$ blocks, $C_{*}$ positive constants.
The $S$ block is a matrix representation of the location of an input pattern relative to a gaze position represented as the $(X/2, Y/2)$th neuron.
The most-activated neuron in the $O$ block indicates the class of the input pattern whose location would be represented by the $S$ block.
The location and identification tasks are obtained by minimizing the first and second terms in the energy function, respectively.
The third and fourth terms represent winner-take-all constraints in the $S$ and $O$ blocks, respectively.

A gaze position is updated by the following rule:
\begin{equation}
l = l + x_{max} - X/2, m = m + y_{max} - Y/2
\label{GazeUpdate}
\end{equation}
where $(x_{max}, y_{max})$ is an index of the most-activated neuron in the $S$ block.
When the gaze position is unchanged by the updating rule, the tasks are complete.

\section{SIMULATION RESULTS}
\begin{figure}[hbtp]
  \begin{center}
    \includegraphics[scale=0.7]{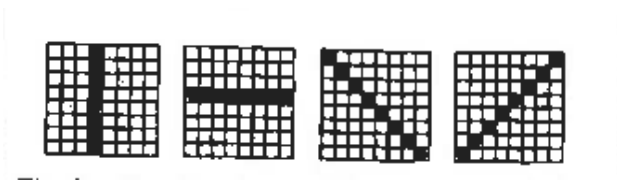}
    \caption{Templates $T^{n}$ used in the first simulation.}
    \label{figure02}
  \end{center}
\end{figure}

\begin{figure}[hbtp]
  \begin{center}
    \includegraphics[scale=0.4]{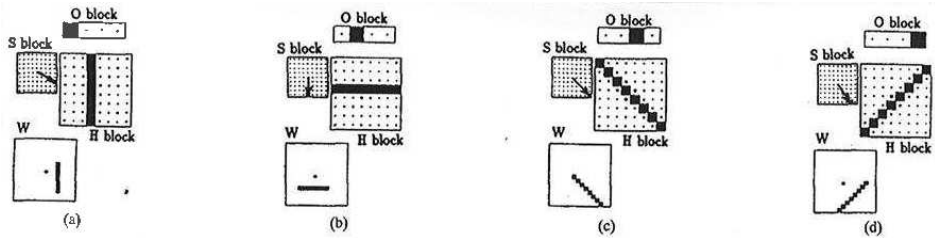}
    \caption{Stable states of the network where the location and the identification task of the line completed.}
    \label{figure03}
  \end{center}
\end{figure}

the first simulation confirmed that the network can perform cooperative subtasks in the saccades.
Positive constants in the energy function and the number of neurons of each block were chosen to be $C_{0} = 0.1$, $C_{1}=0.1$, $C_{2}=1.0$, $C_{3}=1.0$ and $X=Y=9$, $I=J=8$, $N=4$, respectively.
Four templates ($8 \times 8$ binary-valued pixels) are depicted in Fig.\ref{figure02}. 
The stable states of the network are illustrated in Fig.\ref{figure03}(a)-(d), where one side of each square represents an output level of the corresponding neuron, an empty circle in $W$ a gaze position, and a directed line in the S block a saccadic vector.
It is also seen that the activation pattern of the $H$ block resembles the template of the class of the target pattern.

\begin{figure}[hbtp]
  \begin{center}
    \includegraphics[scale=0.7]{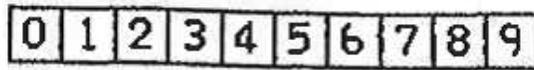}
    \caption{Templates $T^{n}$ used in the second simulation.}
    \label{figure04}
  \end{center}
\end{figure}

\begin{figure}[hbtp]
  \begin{center}
    \includegraphics[scale=0.7]{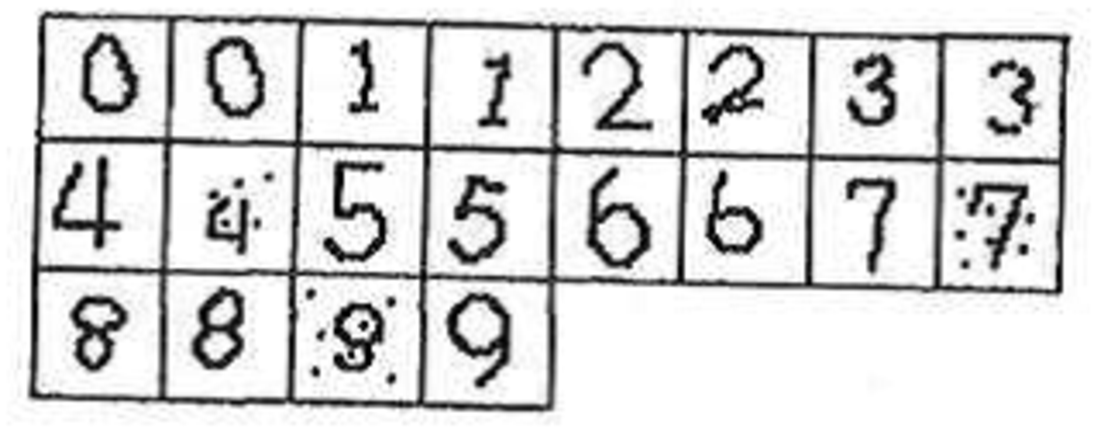}
    \caption{Examples of patterns recognized by the network.}
    \label{figure05}
  \end{center}
\end{figure}

\begin{figure}[hbtp]
  \begin{center}
    \includegraphics[scale=0.7]{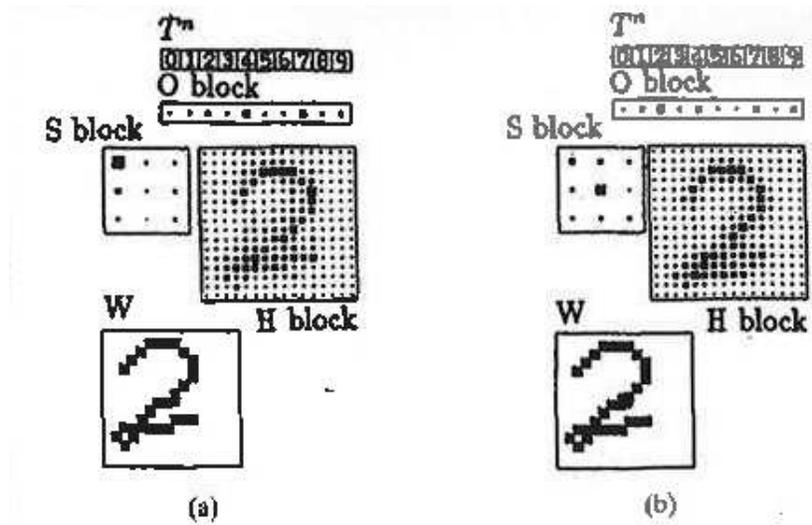}
    \caption{Behavior of the network in the second simulation.}
    \label{figure06}
  \end{center}
\end{figure}

The network was next applied to shift-invariant pattern recognition.
Positive constants in the energy function and the number of neurons of each block were chosen to be $C_{0} = 0.2$, $C_{1}=0.1$, $C_{2}=2.0$, $C_{3}=10.0$ and $X=Y=3$, $I=J=16$, $N=10$, respectively.
Ten templates ($16 \times 16$ binary-valued pixels) are depicted in Fig.\ref{figure04}.
In the experiment, 160 input patterns shifted in any direction by one pixel and deformed by no more than eight hits (Hamming distance), including the patterns shown in Fig.\ref{figure05}, were recognized perfectly.
The behavior of the network is illustrated in Fig.\ref{figure06}.
Figure \ref{figure06}(a) and (b) show the stable states before and after a saccade by one pixel to the upper left, respectively.
We can see the identification task was completed by the saccade that resulted from the location task.

\section{CONCLUSIONS}
A Hopfield neural network for saccades has been proposed.
Computer simulations confirmed that the network performs the location and the identification tasks in a cooperative fashion.
It has also been demonstrated that the network can be applied to shift-invariant pattern recognition.
Since out network is a standard Hopfield-type neural network, it is suitable for VLSI and optical implementations and therefore can speed up saccadic tasks and shift-invariant pattern recognition significantly.

\section{ACKNOWLEDGEMENT}
The author thanks the reviewer who originally recommended the publication of this letter and Dr. Y. Tomita of the University of Electro-Communications for his helpful comments.


\begin{thebibliography}{}
\bibitem{Ballard1989}
D.H. Ballard, "Behavioral constraints on animate vision," Image and Vision Computing, vol.7, No.1 (1989).
\bibitem{Hopfield1982}
J. J. Hopfield, "Neural networks and physical systems with emergent collective computational abilities," Proc.  
Nat. Acad. Sci. U.S., vol.79 (1982).
\bibitem{Nakano:Niizuma:Omori1989}
K. Nakano, M. Niizuma, and T. Omori, "Model of neural visual system with self-organizing cells," Biolog.  
Cybern., vol.60 (1989).
\bibitem{Takeda:Goodman1986}
M. Takeda and J. W. Goodman, "Neural networks for computation: Number representations and programming  
complexity," Appl. Opt., vol.25, no.18 (1986).
\bibitem{Tank:Hopfield1986}
D. W. Tank and J. J. Hopfield, "Simple neural optimization networks: An a/d converter, signal decision circuit,  
and a linear programming circuit," IEEE Trans. Circuits and Syst., vol.CAS-33, no.5 (1986)
\end{thebibliography}
\end{document}